\documentclass[journal]{IEEEtran}
\ifCLASSINFOpdf
\else
\fi
\usepackage{cite}
\usepackage{subfig}
\usepackage{amsmath,amssymb,amsfonts}
\usepackage{dsfont}
\usepackage{algorithmic}
\usepackage{algorithm}
\usepackage{graphicx}
\usepackage{textcomp}
\usepackage{xcolor}
\usepackage{lipsum}
\usepackage{enumitem}
\usepackage{tabularx,booktabs}
\newcolumntype{C}{>{\centering\arraybackslash}X} 
\usepackage{subfig}
\def\BibTeX{{\rm B\kern-.05em{\sc i\kern-.025em b}\kern-.08em
    T\kern-.1667em\lower.7ex\hbox{E}\kern-.125emX}}

\begin{document}
%
\title{A New Look at Spike-Timing-Dependent Plasticity Networks for Spatio-Temporal Feature Learning}
%
%
%

\author{Ali~Safa,~\IEEEmembership{Student Member,~IEEE}, Ilja Ocket, \IEEEmembership{Member, IEEE}, Andr\'e Bourdoux, \IEEEmembership{Senior Member, IEEE}, \\
Hichem Sahli, Francky Catthoor, \IEEEmembership{Fellow, IEEE}, Georges G.E. Gielen, \IEEEmembership{Fellow, IEEE}   
        
\thanks{Ali Safa, Ilja Ocket, Francky Catthoor and Georges G.E Gielen are with imec and the Department of Electrical Engineering, KU Leuven, 3001 Leuven, Belgium (e-mail: Ali.Safa@imec.be; Ilja.Ocket@imec.be; Francky.Catthoor@imec.be; Georges.Gielen@kuleuven.be).}
\thanks{Andr\'e Bourdoux is with imec, 3001 Leuven, Belgium (e-mail: Andre.Bourdoux@imec.be).}
\thanks{Hichem Sahli is with imec and ETRO, VUB, Brussels, Belgium (e-mail: hsahli@etrovub.be).}
}

\maketitle

\begin{abstract}
We present new theoretical foundations for unsupervised Spike-Timing-Dependent Plasticity (STDP) learning in spiking neural networks (SNNs). In contrast to empirical parameter search used in most previous works, we provide novel theoretical grounds for SNN and STDP parameter tuning which considerably reduces design time. Using our generic framework, we propose a class of global, action-based and convolutional SNN-STDP architectures for learning spatio-temporal features from event-based cameras. We assess our methods on the N-MNIST, the CIFAR10-DVS and the IBM DVS128 Gesture datasets, all acquired with a real-world event camera. Using our framework, we report significant improvements in classification accuracy compared to both conventional state-of-the-art event-based feature descriptors (+8.2\% on CIFAR10-DVS), and compared to state-of-the-art STDP-based systems (+9.3\% on N-MNIST, +7.74\% on IBM DVS128 Gesture). Our work contributes to both ultra-low-power learning in neuromorphic edge devices, and towards a biologically-plausible, optimization-based theory of cortical vision. 
\end{abstract}
 
\section{Introduction}
\label{lintro}
\IEEEPARstart{I}{n} recent years, the study of biologically-plausible neural networks has attracted much attention due to the opportunity of implementing them in ultra-low-power and -area \textit{neuromorphic} chips that can learn at the edge.

Unlike conventional deep neural networks (DNNs), a biologically-plausible neural learning architecture complies with three key characteristics. For one, neurons must communicate using binary \textit{action potentials} (also called \textit{spikes}) as opposed to continuous activation functions used in DNNs. Secondly, the inter-neuron communication must be event-driven and asynchronous (vs. frame-based in DNNs) \cite{bookelia}. Finally, network learning must rely on unsupervised \textit{local} learning rules (using the local information to each neuron) such as \textit{Spike-Timing-Dependent Plasticity} (STDP) as empirically observed in the brain \cite{9207239} (vs. conventional error back-propagation and supervised learning in DNNs).

\begin{figure}[htbp]
\centering
    \includegraphics[scale = 0.59]{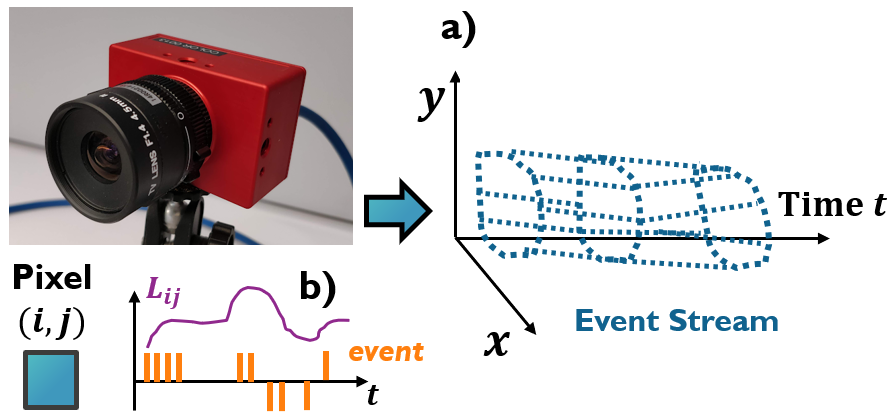}
    \caption{\textit{\textbf{Illustrative example of DVS camera data.} In this conceptual illustration, the DVS is capturing the number "0". a) The camera continuously outputs a stream of events in both space and time. b) When the magnitude change $|\Delta L_{ij}|$ in light log-intensity $L_{ij}$ impinging the pixel $(i,j)$ crosses a threshold $C$, the pixel sends an event. The event is positive if $\Delta L_{ij}>0$ and negative vice versa.}}
    \label{dvsconcept}
\end{figure}

Event-driven cameras (or \textit{Dynamic Vision Sensors}, DVS) are neuromorphic, bio-inspired vision sensors integrating independent pixels $\Bar{p}_{i,j}$ that generate spikes asynchronously whenever the change in light log-intensity $|\Delta L_{i,j}|$ is larger than a threshold $C$ \cite{9138762}. The events are said to have a \textit{positive} polarity when $\Delta L_{i,j}>0$ and vice versa when negative \cite{9138762} (see Fig. \ref{dvsconcept}). 
The advantages of DVS cameras over standard imaging shutters are i) a fine-grain time resolution ($\sim$1$\mu$s) ii) a high dynamic range ($\sim$120dB vs. $\sim$80dB for imaging cameras) and iii) a reduced data communication bandwidth (memory usage) \cite{9138762}. Thanks to those characteristics, DVS cameras are therefore well-suited for IoT applications, where data processing must be done with ultra-low area and power budgets. In addition, the spiking nature of DVS cameras is directly compatible with SNN-STDP systems investigated in this work \cite{9138762}.

The study of biologically-plausible \textit{spiking} neural networks (SNNs) that learn through STDP, coupled with the use of \textit{event-driven vision sensors} will enable a generation of ultra-low-power and -area, computer vision systems \cite{stdplow} integrating unsupervised, on-line learning in the sensor chips themselves, thus reducing latency and preserving the privacy of user data. Therefore, the aim of this work is to propose an SNN-STDP framework for \textit{real-world} event-driven camera data (as opposed to conventional, \textit{static images} converted to spike trains).

Although DVS cameras and SNN-STDP systems fit hand in glove due to their common working principle, reported SNN-STDP architectures are currently less competitive compared to conventional methods. First, they have been demonstrated on a significantly simpler and smaller number of applications, often reaching inferior inference performances (such as classification accuracy) compared to conventional methods \cite{stdplow}. Secondly, designing SNN-STDP systems currently involves the careful hand-tuning of a large number of hyper-parameters (STDP parameters, neuron thresholds, time constants and so on) \cite{stdplow}. Thirdly, it is often unknown which exact \textit{optimization objective} STDP learning is solving, adding uncertainty to their design and behaviour. 

In this paper, we argue that the main limiting factor for the design of high-performance SNN-STDP systems is the lack of a top-down, \textit{optimization-based} theoretical approach that bridges the gap between traditional feature learning and the biologically-plausible SNN setting. Earlier work shed light on the existence of links between unsupervised SNN-STDP architectures and conventional machine learning methods such as PCA, dictionary learning and auto-encoders \cite{lin2018sparse, ae}. Motivated by those earlier works, we provide the following contributions in this paper:
\begin{enumerate}
    \item  We derive a novel theoretical framework for describing SNN-STDP ensembles through the theory of \textit{joint dictionary learning and basis pursuit} (DLBP) via \textit{LASSO} coding \cite{7293682}.
    \item  In contrast to \textit{empirical parameter search} used in most previous works, we derive new theoretical grounds and methods for STDP parameter tuning, network weight initialization, STDP learning termination and neural threshold and time constant tuning.
    \item  we use our methods to engineer SNN-STDP feature descriptors \textit{for DVS data} in both the conventional fully-connected mode and the less-widespread \textit{convolutional} SNN-STDP setting. We assess our framework on three event-based camera datasets for pattern and action recognition: N-MNIST, CIFAR10-DVS and IBM DVS128 Gesture \cite{10.3389/fnins.2015.00437, 10.3389/fnins.2017.00309, 8100264}. 
    \item We report competitive performance against state-of-the-art feature descriptors and SNN-STDP systems ($+9.3 \%$ on N-MNIST vs. prior SNN-STDP works, $+8.2 \%$ on CIFAR10-DVS, $+7.74 \%$ on IBM DVS128 Gesture). 
    \item We provide power consumption estimates to motivate the use of SNN-STDP for edge learning and inference.
\end{enumerate}

This paper is organized as follows. Related works are discussed in Section \ref{related}. Background SNN theory is discussed in Section \ref{background}. Our theoretical SNN-STDP analysis is presented in Section \ref{methodsec}. A study on SNN parameter tuning is provided in Section \ref{paramstud}. The various SNN architectures used in this work are presented in Section \ref{biofeat}. Results are presented in Section \ref{experres}. An SNN power consumption analysis is provided in Section \ref{powercon}. Conclusions are provided in Section \ref{concsec}.

\section{Related Work}
\label{related}
In the past years, a growing number of DVS datasets have been proposed as neuromorphic alternatives to traditional \textit{frame-based} image datasets (such as the standard MNIST). Among them, the \textit{Neuromorphic-MNIST} dataset (N-MNIST) \cite{10.3389/fnins.2015.00437} has been acquired by re-capturing the complete MNIST dataset with a real-world event camera. In addition to N-MNIST, the CIFAR10-DVS has been proposed \cite{10.3389/fnins.2017.00309} following a similar approach than in N-MNIST, by re-capturing the original CIFAR10 images with a  DVS camera. Finally, the IBM DVS128 Gesture dataset \cite{8100264} has been captured for the study of gesture recognition with event-based cameras and features 11 gesture classes. Compared to still images (e.g., standard MNIST or CIFAR10), real-world DVS data adds an additional level of recognition difficulty since real-world event data is rather fuzzy, with a significant amount of spurious spikes \cite{10.3389/fnins.2015.00437}.

Since our aim is to design SNN-STDP architectures for processing \textit{real-world event-based camera data}, we choose to assess our methods on N-MNIST, CIFAR10-DVS and IBM Gesture, and we compare our results to the state-of-the-art performances reported in literature.

In addition to those datasets, a growing number of spatio-temporal feature descriptors have been proposed for summarizing DVS data into vectors that can be classified by conventional techniques such as support vector machines (SVMs) \cite{8578284, 8723171}. Recent among them, the \textit{Histogram of Averaged Time Surfaces} (HATS) \cite{8578284} has been presented as a \textit{hand-crafted} computationally-efficient feature descriptor obtained by concatenating normalized histograms from local patches integrated after a time surface transformation of the DVS data. Complementary to HATS, the \textit{Distribution Aware Retinal Transform} (DART) \cite{8723171} adopts a different method by first aggregating local histograms of events in ellipsoidal log-polar grids, and then applying k-means clustering to learn a codebook from the histograms (used to transform the histograms during inference). It has been demonstrated that DART outperforms HATS on the challenging CIFAR10-DVS dataset while reporting lower performance on N-MNIST \cite{10.3389/fnins.2017.00309}. Therefore, we consider both HATS and DART as state of the art for the class of \textit{non-spiking}, single-layer feature descriptors.

Compared to the aforementioned methods, our framework is biologically plausible and based on a single-layer \textit{spiking} feature descriptor, integrating an SNN with STDP learning. Our method outperforms previous feature descriptors in terms of inference accuracy on both the N-MNIST \cite{10.3389/fnins.2015.00437} and the CIFAR10-DVS \cite{10.3389/fnins.2017.00309} datasets. Our system fully works in the spiking domain and is compatible with ultra-low-power and -area neuromorphic hardware.   
 
In recent years, a number of biologically-inspired SNN-STDP architectures have been proposed in the literature \cite{ae, 8989987, 9206681, doi:10.1126/sciadv.abh0146, chin3}. Recent among them, the MuST framework \cite{8989987} has been presented as the first SNN-STDP architecture incorporating \textit{multi-scale} information in both time and space. Compared to MuST, our architecture achieves a higher accuracy on N-MNIST. Regarding the more challenging 11-class IBM DVS128 Gesture dataset \cite{8100264}, the Tempotron-STDP architecture \cite{chin3}, the STDP-based \textit{reservoir} architecture \cite{9206681} and a hybrid STDP-backprop system \cite{doi:10.1126/sciadv.abh0146} have been proposed. Compared to those works, we report a higher gesture recognition accuracy.  

To the best of our knowledge, this paper presents the first assessment of an SNN-STDP network on the \textit{highly challenging} CIFAR10-DVS dataset (more complex and larger images compared to e.g., N-MNIST digits).
Our theoretical framework take its roots and motivation in several works \cite{lin2018sparse, articleLCA, ae}, but significantly differs from them in terms of architectural differences (e.g., network topology, use of STDP, push-pull spiking neurons and the novel use of an \textit{error layer}).

\section{Background Theory}
\label{background}
\subsection{Leaky Integrate-and-Fire neuron}
\begin{figure}[htbp]
\centering
    \includegraphics[scale = 0.55]{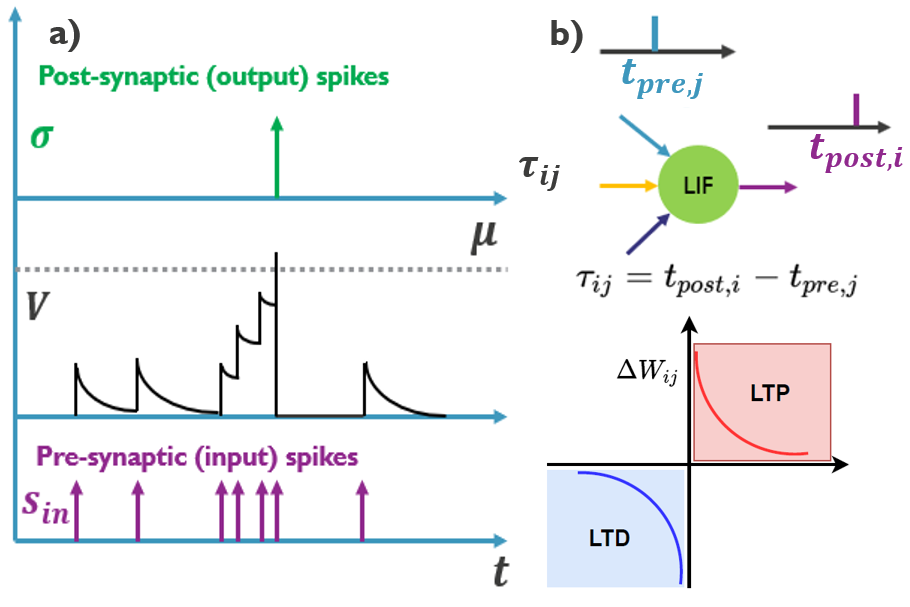}
    \caption{\textit{\textbf{Conceptual illustration of the LIF neuron and STDP learning}. a) The LIF neuron is connected to a pre-synaptic spiking input $s_{in}$. When the membrane potential $V$ crosses the threshold $\mu$, the neuron emits a spike $\sigma = 1$ and $V$ is reset to zero. b) The weight $W_{ij}$ is modified according to the double exponential STDP rule (\ref{stdppp}) in function of the difference between the pre- and the post-synaptic spike times $\tau_{ij} = t_{post,i} - t_{pre,j}$.}}
    \label{lifconcept}
\end{figure}

In contrast to traditional DNNs, SNNs make use of \textit{spiking} neurons as non-linearity, often modelled by a Leaky Integrate-and-Fire (LIF) activation:
\begin{equation}
 \begin{cases}
    \frac{dV}{dt} = \frac{1}{\tau_m} (J_{in} - V)
    \\
    \sigma = 1 \hspace{3pt} \text{\textbf{if}} \hspace{3pt} V \geq \mu \hspace{3pt} \text{\textbf{else}} \hspace{3pt} 0
    \\
    V(t+dt) = 0 \hspace{3pt} \text{\textbf{if}} \hspace{3pt} V \geq \mu
  \end{cases}
  \label{liff}
\end{equation}
with $J_{in}$ the input current to the neuron, $\sigma$ the spiking output, $V$ the membrane potential, $\tau_m$ the time constant governing the membrane potential decay, $dt$ is the simulation time step and $\mu$ the neuron threshold \cite{bookelia}. The scalar input current $J_{in}$ is continuously integrated in $V$ following (\ref{liff}). When $V$ crosses the firing threshold $\mu$, the membrane potential is reset back to zero and an output spike is emitted. The input current $J_{in}$ is obtained by filtering the inner product of the neural weights and the spiking inputs through a \textit{post-synaptic current} (PSC) kernel \cite{bookelia} (estimating the spiking \textit{rate}):
\begin{equation}
    J_{in} = \mathcal{PSC}\{ \Bar{\phi}^T \Bar{s}_{in}(t)  \}
    \label{inee}
\end{equation}
with $\Bar{s}_{in}(t)$ the input spiking vector (originating from e.g. an event camera or other spiking neurons), $\Bar{\phi}$ the weight vector and:
\begin{equation}
    \mathcal{P}\mathcal{S}\mathcal{C} \{ \theta(t) \} = \theta(t) * \frac{1}{\tau_s} e^{-t/\tau_s}
\end{equation}
the effect of PSC filtering with time constant $\tau_s$. Fig. \ref{lifconcept} a) conceptually illustrates the LIF neuron behaviour.


\subsection{STDP learning}
\label{weargue}
Even though back-propagation training using surrogate gradients has shown remarkable success in conventional GPU-based supervised training settings \cite{alirad}, its use is still ill-suited for ultra-low power edge learning as it suffers from the same compute-expensive credit assignments issues faced in standard DNN back-propagation \cite{8528875}. In addition, the surrogate gradient method is non-local and hence, biological implausible, making it ill-suited for the study and modelling of cortical neural ensembles. Therefore, the study of Spike-Timing-Dependent Plasticity (STDP) is actively being pursued as a biologically plausible and local learning rule, avoiding the compute-expensive credit assignment problem by \textit{locally} modifying the weights of each neuron following the difference $\tau_{ij}$ between the post- and pre-synaptic spike times \cite{Bi10464}: 
\begin{equation}
   W_{ij} \xleftarrow{} \begin{cases}
  W_{ij} + A_{+} e^{-\tau_{ij}/ \tau_+}, & \text{if } \tau_{ij} \geq 0
\\
   W_{ij} -A_{-} e^{\tau_{ij}/ \tau_-}, & \text{if } \tau_{ij} < 0
\end{cases}
\label{stdppp}
\end{equation}
with $A_+, A_-$ the potentiation and depression weights, $\tau_+, \tau_-$ the potentiation and depression decay constants, $W_{ij}$ the $j^{th}$ element of the $i^{th}$ neuron weight vector $\bar{W}$ and $\tau_{ij}$ the time difference between the post- and the pre-synaptic spike times across the $j^{th}$ synapse of neuron $i$ (see Fig. \ref{lifconcept} b) \cite{Bi10464}. 
 
On the other hand, a wider adoption of the STDP learning rule is still lagging far behind the use of surrogate gradient back-propagation due to significant gaps in reported performances and task complexity compared to the latter. We argue that these shortcomings are mostly due to the lack of a comprehensive optimisation-based framework linking SNN-STDP systems to conventional unsupervised dictionary learning algorithms. Indeed, the design of most SNN-STDP systems is highly empirical and typically consists of randomly initialising a network topology composed of excitatory and inhibitory neurons (i.e., \textit{positive} or \textit{negative} contribution to the post-synaptic input current). The network topology is randomly initialised in a way such that competition is created between the excitatory neurons by preventing inhibitory neurons to input the same excitatory neurons that they receive as inputs. Then, STDP learning is added to the excitatory neurons. Such system is therefore composed of a large number of parameters (neuron threshold, time constants, connection probabilities, STDP parameters and so on) that are typically hand-tuned through a lengthy and unsystematic process. Finally, since such random topology cannot be written under a specific loss function, usual cross-validation techniques cannot be directly used to detect when over-fitting starts to occur in order to stop training.

Therefore, in order to consolidate an optimisation-based description of STDP learning, we analyse in Section \ref{methods} the SNN-STDP problem under the light of joint dictionary learning and basis pursuit (DLBP).
\begin{figure*}[!t]
\centering
\captionsetup{justification=centering}
    \includegraphics[scale = 0.52]{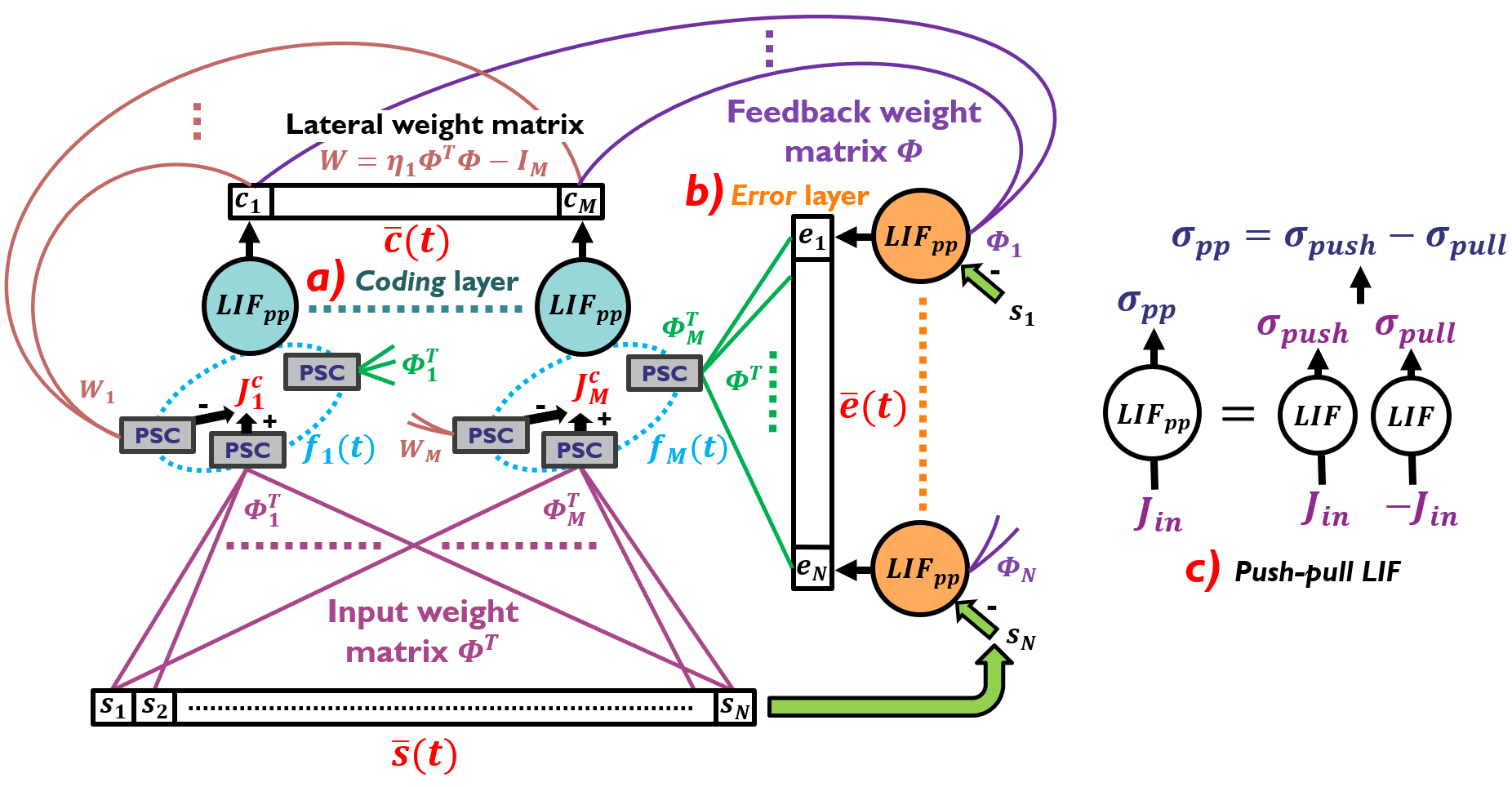}
    \caption{\textit{\textbf{Proposed SNN-STDP architecture.} a) The coding layer is composed of $i=1,...,M$ push-pull LIF neuron pairs $LIF_{pp}$ receiving the input data $\bar{s}$ through the weight matrix $\Phi^T$ and the \textbf{past} output vector $\bar{c}$ through the lateral weight matrix $W$. Using those two input sources, the coding layer can perform LASSO coding as demonstrated in Section \ref{baspurs}. The coding layer also receives the re-projection error vector $\bar{e}$ through the weight matrix $\Phi^T$ and combines it with the other inputs following (\ref{waytocompute}) to obtain the spike train $f_i$ for each coding neuron. During STDP learning of $\Phi^T$ and $W$, $f_i$, $c_i$ are used as post-synaptic and $\bar{e}$ as pre-synaptic spike trains. b) The error layer is composed of $j = 1,...,N$ push-pull LIF neuron pairs $LIF_{pp}$ receiving as input the entry $s_j$ and the network output $\bar{c}$ through the feedback weight matrix $\Phi$ (remains identical to the transposed of input weights during learning). The error layer outputs the error vector $\bar{e}$. During STDP learning of $\Phi$, each $e_j(t)$ is used as the post-synaptic spike train and $\bar{c}$ is used as the pre-synaptic spike trains. c) The push-pull LIF neuron can code both positive and negative values of $J_{in}$, and output spikes of positive or negative polarity.}}
    \label{fig1}
\end{figure*}

\section{Methods}
 \label{methods}
 \label{methodsec}
\subsection{SNN-STDP topology}

Our proposed SNN-STDP topology is shown in Fig. \ref{fig1} and solves the DLBP problem (\ref{lassodefdivc}) thanks to its specific wiring architecture.
\begin{equation}
    \Bar{c}, \Phi = \arg \min_{\Bar{c}, \Phi} \frac{1}{2} ||\Phi \Bar{c} - \Bar{s}||_{2}^{2} + \lambda_1 ||\Bar{c}||_1 + \frac{\lambda_2}{2} ||\Phi||_F^2
    \label{lassodefdivc}
\end{equation}
where $\Phi$ is the network weight matrix, $\bar{c}$ is the network output, $\bar{s}$ is the input to the network, $\lambda_1$ is a sparsity defining hyper-parameter and $\lambda_2$ is a weight decay term.

The proposed SNN-STDP system solves (\ref{lassodefdivc}) by \textit{i)} inferring the next output code $\Bar{c}$ using the current weights $\Phi$ and \textit{ii)} learning the next weights $\Phi$ using the current output $\bar{c}$, by iterating along the time dimension of the event-based data. During the training phase, we feed event-based acquisitions one by one to the network. Section \ref{baspurs} and \ref{dicolea} respectively demonstrate how \textit{i)} and \textit{ii)} are jointly solved using the proposed architecture. In addition, Fig. \ref{fig1} c) the neural unit $LIF_{pp}$ used throughout this work and constructed using a pair of push-pull LIF neurons. Each neuron $LIF_{pp}$ possesses its own weight vector which is modified locally during learning by the STDP mechanism. Our use of $LIF_{pp}$ as neural units (i.e., positive and negative outputs) allows the network to perform an approximate LASSO regression and is biologically motivated by empirical evidence about the existence of push-pull neural pairs in the cortex \cite{10.2307/2881116} and in retinal ganglions used to code the light intensities \cite{jphysiol.1962.sp006837}. 

The coding neurons (Fig. \ref{fig1} a) are wired following:
\begin{equation}
    \bar{J}_{in}^c = \mathcal{PSC} \{ \eta_1 \Phi^T \Bar{s} - (\eta_1 \Phi^T \Phi - I_M)\Bar{c} \}
    \label{lautre}
\end{equation}
and solves the \textit{LASSO} sub-problem \textit{i)} of (\ref{lassodefdivc}) as shown in Section \ref{baspurs}. In addition, the input to the \textit{error} neurons (Fig. \ref{fig1} b) is wired following:
\begin{equation}
    \bar{J}_{in}^e = \mathcal{PSC} \{ \Phi \Bar{c}-\Bar{s} \}
    \label{errorwiring}
\end{equation}
and the error output $\Bar{e}(t)$ is wired back to the \textit{coding} neurons through a \textit{separate} set of weights $\Phi^T$ (as in the \textit{pyramidal neurons} of the visual cortex  which posses multiple sets of synapses \cite{NEURIPS2018_1dc3a89d}). Such topology solves the \textit{dictionary learning} sub-problem \textit{ii)} of (\ref{lassodefdivc}) when STDP is applied on all weights, as demonstrated in Section \ref{dicolea}.

\subsection{LASSO Basis Pursuit}
\label{baspurs}
Now, we show how the coding neurons of our SNN topology in Fig. \ref{fig1} a) solves the LASSO basis pursuit part of (\ref{lassodefdivc}).
Considering a fixed $N\times M$ dictionary $\Phi$ featuring $M$ atoms, the LASSO regression is written as \cite{cpa.20042}:
\begin{equation}
    \Bar{c} = \arg \min_{\Bar{c}} \frac{1}{2} ||\Phi \Bar{c} - \Bar{s}||_{2}^{2} + \lambda_1 ||\Bar{c}||_1
    \label{lassodef}
\end{equation}
with $\Bar{c}$ the $M$-dimensional sparse code, $\Bar{s}$ the $N$-dimensional input vector and $\lambda_1$ a hyper-parameter defining the number of non-zero entries (or \textit{support}) $\Theta$ of $\Bar{c}$ (the higher $\lambda_1$, the smaller $\Theta$). The pursuit (\ref{lassodef}) is classically solved using the \textit{iterative soft-thresholding algorithm} (ISTA) \cite{cpa.20042}:
\begin{equation}
    \Bar{c}^{k+1} = \text{Prox}_{\lambda_1 ||.||_1}\{\Bar{c}^{k} - \eta_1 \Phi^T \Phi \Bar{c}^{k} + \eta_1 \Phi^T \Bar{s} \}
    \label{ista}
\end{equation}
with $\eta_1$ the coding rate, $k$ the iteration index and 
\begin{equation}
    \text{Prox}_{\lambda_1 ||.||_1}\{ \Bar{x}\}_i = \text{sign}(x_i) \max(|x_i| - \lambda_1, 0)
    \label{proxi}
\end{equation}
Let $\bar{J}_{in}^c$ the input current vector to the coding neurons defined in (\ref{lautre}) following the argument of $\text{Prox}_{\lambda_1 ||.||_1}$ in (\ref{ista}). 

\begin{figure}[htbp]
\centering
    \includegraphics[scale = 0.45]{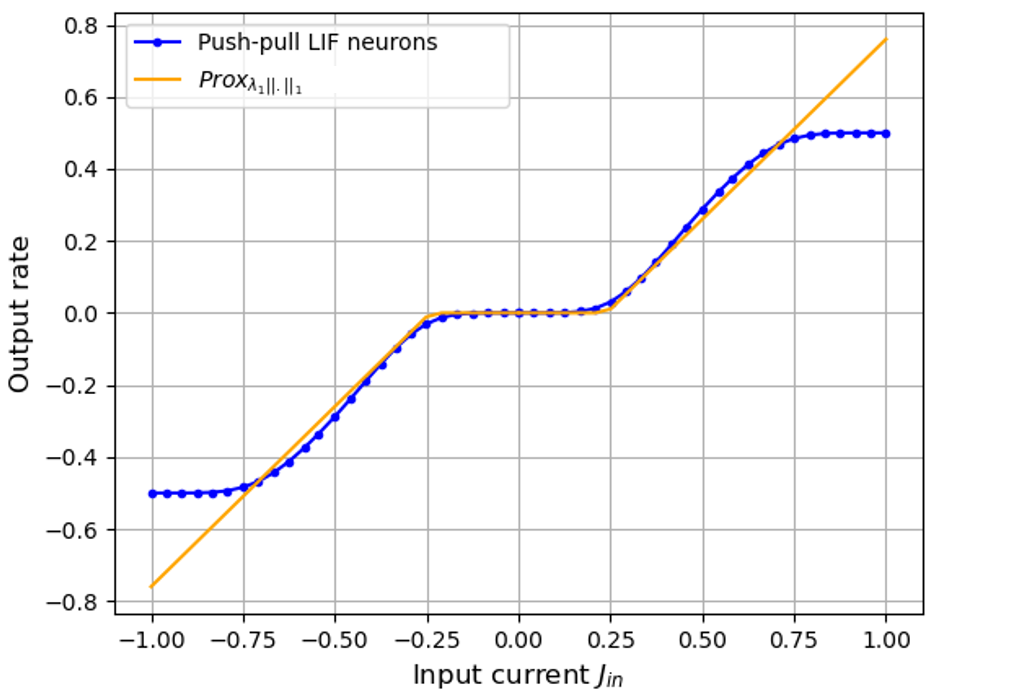}
    \caption{\textit{\textbf{Simulation of a push-pull pair of LIF neurons} approximating the proximal operator (\ref{proxi}) in the non-saturating region of the output spike rate. The neuron threshold $\mu$ is set to 0.25 in this example and the constraint (\ref{cond1}) is used to set $\tau_m$. The weight decay term $\frac{\lambda_2}{2} ||\Phi||_F^2$ in (\ref{lassodefdivc}) helps keeping the spike rates within the non-saturating region.}}
    \label{proxiimg}
\end{figure}

We define the $i^{th}$ pair of LIF neurons as follows \cite{8888024}:
\begin{equation}
\frac{d}{dt}
\begin{bmatrix}
V_{p}^i\\
V_{n}^i
\end{bmatrix}=
\frac{1}{\tau_m} (
\begin{bmatrix}
J_{in, i}^c \\
-J_{in, i}^c 
\end{bmatrix} - 
\begin{bmatrix}
V_{p}^i\\
V_{n}^i
\end{bmatrix}
)
\label{membranepot}
\end{equation}
with $V_p^i$ and $V_n^i$ respectively denoting the $i^{th}$ push or pull membrane potential and $\tau_m$ denoting the membrane decay constant. Whenever $V_p^i$ or $V_n^i$ exceeds the neuron threshold $\mu$, the corresponding push or pull neuron outputs a spike and the corresponding membrane potential is reset to $0$:
\begin{equation}
\begin{bmatrix}
\sigma_{p}^i\\
\sigma_{n}^i
\end{bmatrix} 
= \mathds{1} \{
\begin{bmatrix}
V_{p}^i \geq \mu\\
V_{n}^i \geq \mu
\end{bmatrix}
\}
\label{indic}
\end{equation}
where $\mathds{1}\{.\}$ is the indicator function (equal to 1 if the condition holds true and 0 otherwise) and $\sigma_{p}^i$ and $\sigma_{n}^i$ respectively  denote the $i^{th}$ push or pull output. 
In order to show how (\ref{membranepot}) solves (\ref{lassodef}), it follows from the LIF spike rate domain analysis in \cite{bookelia} that a first-order Taylor approximation around $\mu$ between the input $J_{in, i}$ and the push-pull spiking rates $r_{p,n}^i$ of the $i^{th}$ \textit{coding} neuron is given by:
\begin{equation}
\begin{bmatrix}
r_{p}^i\\
r_{n}^i
\end{bmatrix}
\approx
\frac{1}{\tau_m \mu}
\begin{bmatrix}
J_{in, i}^c - \mu \hspace{3pt} \text{\textbf{if}} \hspace{3pt} J_{in, i}^c > \mu \hspace{3pt} \text{\textbf{else}} \hspace{3pt} 0\\
J_{in, i}^c + \mu \hspace{3pt} \text{\textbf{if}} \hspace{3pt} J_{in, i}^c < -\mu \hspace{3pt} \text{\textbf{else}} \hspace{3pt} 0
\end{bmatrix}
\label{taylort}    
\end{equation}
which is equivalent to the proximal (\ref{proxi}) if:
\begin{equation}
    \tau_m = \frac{1}{\mu}
    \label{cond1}
\end{equation}
in order to enforce a unit slope (see Fig. \ref{proxiimg}). 
Therefore, a layer containing $M$ \textit{coding} neurons acts as (\ref{proxi}) in the spiking rate sense. Using (\ref{taylort}) and (\ref{cond1}), the effect of the coding layer (in Fig. \ref{fig1} a) of our SNN-STDP system can be written as: 
\begin{equation}
\Bar{c}(t) = 
\begin{bmatrix}
1 & -1
\end{bmatrix}
\begin{bmatrix}
\Bar{r}_{p}\\
\Bar{r}_{n}
\end{bmatrix} \approx 
\text{Prox}_{\mu ||.||_1}\{
\bar{J}_{in}^c
\}
\label{membranepotmodif}
\end{equation}
Finally, it directly results from (\ref{ista}) that wiring the coding neurons as follows:
\begin{equation}
    \bar{J}_{in}^c = \mathcal{P}\mathcal{S}\mathcal{C} \{\eta_1 \Phi^T \Bar{s}(t) - W\Bar{c}(t)\}
    \label{setting}
\end{equation}
with:
\begin{equation}
W \equiv \eta_1 \Phi^T \Phi - I_M
\end{equation}
makes the SNN behave as (\ref{ista}) during the iterations which in turn, makes the push-pull output spiking rates converge to the solution of (\ref{lassodef}). 

\subsection{Learning the network weights}
\label{dicolea}
At this point, we have shown how our SNN architecture solves LASSO. We still need to show how the SNN topology in Fig. \ref{fig1} solves the dictionary learning part of (\ref{lassodefdivc}).
Let $\bar{c}$ be fixed, the dictionary learning problem can be written as \cite{8237828}:
\begin{equation}
    \Phi = \arg \min_{\Phi} \frac{1}{2} ||\Phi \Bar{c} - \Bar{s}||_{2}^{2} + \frac{\lambda_2}{2} ||\Phi||_F^2
    \label{dicodef}
\end{equation}
with $\lambda_2$ a weight decay hyper-parameter (often referred to as \textit{homeostasis} in neuroscience \cite{10.3389/fncom.2019.00049}). Eq. (\ref{dicodef}) is traditionally solved by gradient descent:
\begin{equation}
    \Phi^{k+1} = \Phi^k - \eta_2(\Phi^k \Bar{c} - \Bar{s})\Bar{c}^T - \eta_2 \lambda_2 \Phi^k
    \label{iterdico}
\end{equation}
where $\eta_2$ defines the learning speed. The rest of this section shows how (\ref{dicodef}) is solved by our SNN-STDP system. 
\subsection{A rate-based STDP model}
\begin{figure}[!t]
\centering
    \includegraphics[scale = 0.57]{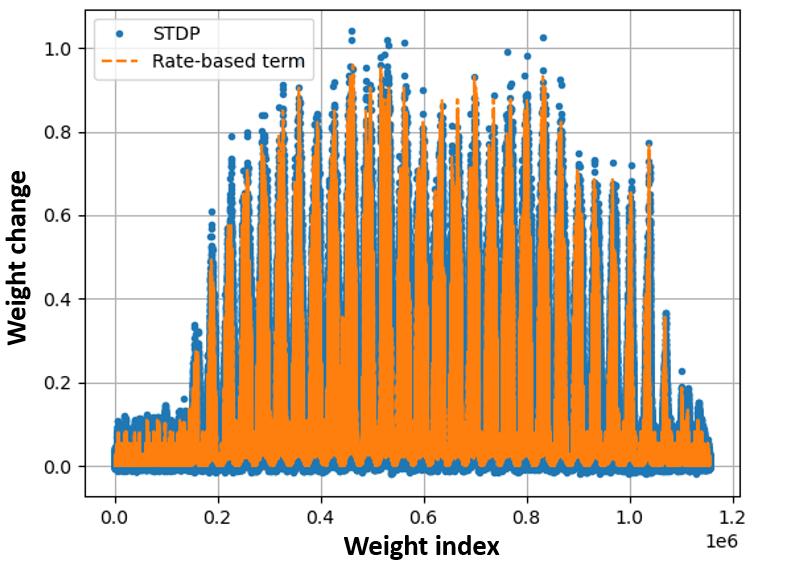}
    \caption{\textit{\textbf{Synaptic modification due to STDP vs. rate-only weight change} in (\ref{approx}), obtained by recording the accumulated weight change of the input synapses to the coding layer $\Phi$ for 300 time-steps using the N-MNIST dataset \cite{10.3389/fnins.2015.00437}. The error of the rate-only model is $ \frac{1}{MN}\sum_{i = 1}^{MN} |\Delta w_{\text{STDP}}^i - \Delta w_{\text{rate}}^i| \approx 2\times 10^{-5}$.}}
    \label{blockdia}
\end{figure}

We note $\tau = t_{post} - t_{pre}$ the time difference between a post- and a pre-synaptic spike. The weight strength is modified following the classical double exponential STDP rule \cite{Bi10464}:
\begin{equation}
    \kappa(\tau) = \begin{cases}
  A_{+} e^{-\tau/ \tau_+}, & \text{if } \tau \geq 0
\\
   -A_{-} e^{\tau/ \tau_-}, & \text{if } \tau < 0
\end{cases}
\label{stdp}
\end{equation}
with $\tau_+$, $\tau_-$ the potentiation and depression time constants and $A_+$, $A_-$ the potentiation and depression weights. Given a pre- and a post-synaptic spike train $s_{pre}(t)$, $s_{post}(t)$, the weight modification of the synapse at the post-synaptic time instant $t$ can be written as \cite{10.1371/journal.pcbi.1007835}:
\begin{equation}
    \Delta w|_t = \eta_2 \int_{-\infty}^{\infty} s_{post}(t) s_{pre}(t - \tau) \kappa(\tau) d\tau
    \label{baseinteg}
\end{equation}
Under this form, (\ref{baseinteg}) is not useful since the post- and pre-synaptic spike trains are not known beforehand. 
Taking the expected value of (\ref{baseinteg}) over $t$ leads to:
\begin{equation}
    \mathcal{E} \{ \Delta w|_t \} = \eta_2 \int_{-\infty}^{\infty} \mathcal{E} \{s_{post}(t) s_{pre}(t - \tau) \} \kappa(\tau) d\tau
    \label{expected}
\end{equation}
the right-hand side of (\ref{expected}) can be re-written as:
\begin{equation}
    \mathcal{E}\{ \Delta w|_t \} = \eta_2 \int_{-\infty}^{\infty} \{ r_{post} r_{pre} + \mathcal{C}(\tau) \}  \kappa(\tau) d\tau 
    \label{expected2}
\end{equation}
where $\mathcal{C}(\tau)$ is the covariance between the pre- and post-synaptic spike train and $r_{post}, r_{pre}$ are the average post- and pre-synaptic piking rates. We further separate (\ref{expected2}) into a rate-only term and a covariance term:
\begin{multline}
    \Delta w_{\text{STDP}} \equiv \mathcal{E}\{ \Delta w|_t \} = \eta_2 (A_+ \tau_+ - A_- \tau_-)r_{post} r_{pre} \\ + \eta_2 \int_{-\infty}^{\infty}  \mathcal{C}(\tau)  \kappa(\tau) d\tau  \hspace{20pt} 
    \label{expected3}
\end{multline}
At this point, it is interesting to evaluate the contribution of each term on synaptic plasticity. Fig. \ref{blockdia} experimentally compares the weight change induced by STDP $\Delta w_{\text{STDP}}$ (\ref{expected3}) against the weight change due to the rate-based term only $\Delta w_{\text{rate}}=\eta_2 (A_+ \tau_+ - A_- \tau_-)r_{post} r_{pre}$. 
It can be clearly remarked in Fig. \ref{blockdia} that the rate-based term of (\ref{expected3}) is the main driver of the \textit{long-term} STDP weight change. Therefore, we approximate $\Delta w_{\text{STDP}}$ as follows in the rest of our derivations: 
\begin{equation}
    \Delta w_{\text{STDP}} \{ s_{post}, s_{pre}\} \approx \eta_2 (A_+ \tau_+ - A_- \tau_-)r_{post} r_{pre}
    \label{approx}
\end{equation}
where it is assumed (without loss of generality) that:
\begin{equation}
    A_+ \tau_+ - A_- \tau_- = 1
    \label{conts1}
\end{equation}
in order to not modify the resulting learning speed (setting $A_+ \tau_+ - A_- \tau_->0$ is enough for the learning effect to take place). Using (\ref{approx}), it is now possible to show how the SNN weights (input weights, feedback weights and lateral weights) can learn the dictionary $\Phi$.

\subsection{Using STDP for dictionary learning}
\subsubsection{Input weights}
\label{inputsec}
A \textit{coding} neuron has in its possession one row $i$ of the input weight matrix $\Phi^T$, noted $\Phi_i^T$. Each element $j$ of $\Phi_i^T$ is modified using the local STDP update rule with homeostasis regularization:
\begin{equation}
    (\Phi_i^T)_j \xleftarrow{} (\Phi_i^T)_j - \Delta w_{\text{STDP}} \{c_i(t), e_j(t)\} - \eta_2 \lambda_2 (\Phi_i^T)_j
    \label{firstst1}
\end{equation}

By using (\ref{approx}) and (\ref{conts1}), we can re-write (\ref{firstst1}) as:
\begin{equation}
    (\Phi_i^T)_j \xleftarrow{} (\Phi_i^T)_j - \eta_2 r \{c_i(t) \} r \{ e_j(t) \}  - \eta_2 \lambda_2 (\Phi_i^T)_j
    \label{firstst12223}
\end{equation}
where $r\{.\}$ denotes the spike rate (similar to $r_{post}$, $r_{pre}$ used earlier). Since our SNN architecture in Fig. \ref{fig1} ensures: 
\begin{equation}
     r\{e_j \} = r\{ \Phi_j \Bar{c} - s_j \}
    \label{rrr}
\end{equation}
it can be clearly seen that (\ref{firstst1}) is solving (\ref{iterdico}) under the STDP rate-based approximation (\ref{approx}).
\\

\subsubsection{Feedback weights}
An \textit{error} neuron has in its possession one row $j$ of the input weight matrix $\Phi$, noted $\Phi_j$. Each element $i$ of $\Phi_j$ is modified using the local STDP update rule with homeostasis regularisation:
\begin{equation}
    (\Phi_j)_i \xleftarrow{} (\Phi_j)_i - \Delta w_{\text{STDP}} \{e_j(t), c_i(t)\} - \eta_2 \lambda_2 (\Phi_j)_i
    \label{firstst2}
\end{equation}
it follows directly from the earlier discussion on input synapse modification (section \ref{inputsec}) that (\ref{firstst2}) is indeed solving (\ref{iterdico}) for the weights in the error layer.
\\

\subsubsection{Lateral weights}
In order to learn the lateral weights $W$, we adapt the \textit{consistency-enforcing} method proposed in \cite{lin2018sparse}:
\begin{equation}
\Tilde{W} = \arg \min_{\Tilde{W}} \frac{1}{2} ||(\Tilde{W} - \Phi^T \Phi ) \Bar{c}||_2^2 
\label{catchup}
\end{equation}
 to our custom use of the STDP rule, where:
 \begin{equation}
     W \equiv \eta_1 \Tilde{W} - I_M
 \end{equation}
Eq. (\ref{catchup}) is traditionally solved by gradient descent as:
\begin{equation}
    \Tilde{W}^{k+1} = \Tilde{W}^{k} - (\Tilde{W}^{k} - \Phi^T \Phi) \Bar{c} \Bar{c}^T
    \label{cathciter}
\end{equation}
Each \textit{coding} neuron $i$ only posses the $i^{th}$ row of $W$ and $\Phi^T$. In addition, the coding neuron $i$ receives as inputs the spike train vectors $\Bar{s}(t)$, $\Bar{c}(t)$ and $\Bar{e}(t)$. It is possible to \textit{locally} compute the internal state $f_i(t)$ representing $(\Tilde{W} - \Phi^T \Phi)_i \Bar{c}$ for each coding neuron $i$:
\begin{multline}
   f_i(t) =   \frac{1}{\eta_1}(W + I_M)_i \Bar{c}(t) - (\Phi^T)_i \Bar{e}(t) - (\Phi^T)_i \Bar{s}(t)  \\ 
   = (\Tilde{W} - \Phi^T \Phi)_i \Bar{c}(t) 
   \label{waytocompute}
\end{multline}
since the wiring (\ref{errorwiring}) in our SNN topology ensures:
\begin{equation}
    r \{\Bar{e}(t) + \Bar{s}(t) \} = r \{ \Phi \Bar{c}(t)\}
\end{equation}
it is now clear that the iterative STDP updates (\ref{firstst121}) (with $i=1,..,M$ spanning the rows and $l=1,...,M$ spanning the columns of $\Tilde{W}$) solve (\ref{catchup}, \ref{cathciter}). 
\begin{equation}
    (\Tilde{W}_i)_l \xleftarrow{} (\Tilde{W}_i)_l - \Delta w_{\text{STDP}} \{f_i(t), c_l(t)\} - \eta_2 \lambda_2 (\Tilde{W}_i)_l
    \label{firstst121}
\end{equation}

Therefore, the above derivations have shown that, under the rate-based assumption, our proposed SNN-STDP network jointly solves the basis pursuit (\ref{lassodef}) and the dictionary learning (\ref{dicodef}) problems. In Section \ref{paramstud}, the following remaining questions will be addressed:
\begin{enumerate}
    \item How to tune the parameters of the STDP rule (\ref{stdp})?
    \item How to initialize the network weights?
    \item When to terminate the STDP training process?
    \item How to optimally tune the LIF neuron threshold $\mu$ and time constant $\tau_m$ in (\ref{membranepot}, \ref{indic})?
\end{enumerate}

\begin{figure}[htbp]
\centering
    \includegraphics[scale = 0.63]{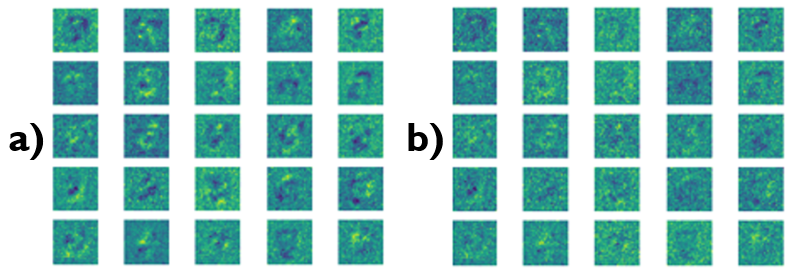}
    \caption{\textit{\textbf{SNN weights learnt on N-MNIST.} a) A matching STDP kernel leads to crisp atoms, mostly capturing feature signals b) A non-matching STDP kernel leads to fuzzy atoms containing a significant amount of noise (details of images better visualized in .pdf rendering than in printed format). }}
    \label{badgood}
\end{figure}


\section{Parameter Study}
\label{paramstud}
\subsection{STDP parameter tuning}
\begin{figure}[!t]
\centering
    \includegraphics[scale = 0.57]{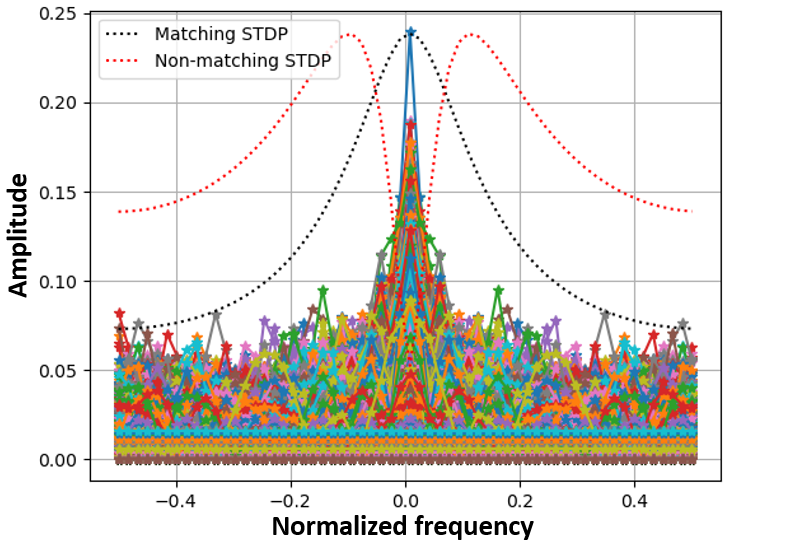}
    \caption{\textit{\textbf{Spectra from event-based camera data} obtained by applying the DFT (with Hanning windowing in order to reduce spectral leakage \cite{1455106}) on spike train of each pixel from the N-MNIST dataset \cite{10.3389/fnins.2015.00437}. In order to obtain a \textit{matching} STDP kernel, $\omega_2$ is tuned such that the -3dB point covers the information-bearing band (with the example parameters $A_+=1, A_-=0.8, \tau_+ = 0.0208 \text{s}, \tau_- = 0.008 \text{s}$). On the other hand, the non-matching STDP kernel ($A_+=1, A_-=0.8, \tau_+ = 0.008 \text{s}, \tau_- = 0.008 \text{s}$) does not comply with the constraint (\ref{canc}) and degrades the information-bearing band of the event data.}}
    \label{spectrum}
\end{figure}
Eq. (\ref{baseinteg}) indicates that STDP acts as a kernel $\kappa(\tau)$ filtering $s_{pre}(t)$ over a certain bandwidth $B \{ A_{+}, A_{-}, \tau_{+}, \tau_{-} \}$. Therefore, it must be ensured that the pass-band is matching the information-bearing part of the $s_{pre}(t)$ spectrum while filtering noise and spurious frequencies caused by distortion (due to the discontinuous nature of spike trains). By taking the Fourier transform of (\ref{stdp}), we obtain:
\begin{equation}
    \mathcal{F}\{\kappa(t) \} = \frac{A_{+}}{\tau_{+}^{-1} + j\omega} - \frac{A_{-}}{\tau_{-}^{-1} - j\omega}
    \label{four1}
\end{equation}
Taking the magnitude of (\ref{four1}), we obtain:
\begin{equation}
    |\mathcal{F}\{\kappa(t) \}| = \frac{A_{+} \sqrt{(\tau_-^{-1} - \alpha \tau_+^{-1})^2 + (1+\alpha)^2\omega^2}}{\sqrt{\tau_+^2 + \omega^2}. \sqrt{\tau_-^2 + \omega^2}}
    \label{spectralmag}
\end{equation}
with $\alpha = A_- / A_+$. When analysing (\ref{spectralmag}), two poles $\omega_{1,2}$ and one zero $z_1$ can be identified:
\begin{equation}
    \omega_1 = \tau_+^{-1} \hspace{10pt} \omega_2 = \tau_-^{-1} \hspace{10pt} z_1 = \frac{\tau_-^{-1} - \alpha \tau_+^{-1}}{1 + \alpha}
    \label{collpole}
\end{equation}

Therefore, (\ref{baseinteg}, \ref{collpole}) show that the double exponential STDP rule acts as a first-order bandpass filter applied on $s_{pre}(t)$. 
Under the Poisson rate-based assumption of this work, the information carried by a spike train $s(t)$ is considered to be its \textit{mean} spiking rate and therefore, solely resides in the DC bin $\mathcal{F} \{ s(t)\}|_{\omega = 0}$ (all other bins containing distortion and noise). Consequently, the STDP kernel must avoid cutting the DC frequency while attenuating AC bins.
Regarding real-world event-based camera data, the spike train bandwidth covers a narrow information band around DC (see Fig. \ref{spectrum}). Consequently, the same discussion to the rate-based Poisson case holds (i.e., STDP must not cut the spectra near DC). 
By pole-zero cancellation of $\omega_1$ and $z_1$ in (\ref{spectralmag}), (\ref{collpole}), a nearly flat STDP spectrum around DC can be obtained, leading to the constraint:
\begin{equation}
    \tau_+^{-1} =  \frac{\tau_-^{-1} - \alpha \tau_+^{-1}}{1 + \alpha}
    \label{canc}
\end{equation}
Fig. \ref{spectrum} shows both a matching STDP kernel, complying with (\ref{canc}), and a non-matching one. 
The non-matching STDP spectrum significantly attenuates the information-bearing bands while passing noise (see Fig. \ref{spectrum}). Therefore, we expect the non-matching STDP kernel to lead to the learning of \textit{noisier} atoms $\Phi_i$ compared to the matching STDP case, degrading the overall quality of the learned dictionary. Fig \ref{badgood} experimentally confirm this discussion and the use of the constraint (\ref{canc}) when setting the STDP parameters (we used the matching and non-matching parameters of Fig. \ref{spectrum} and kept all other learning parameters constant during the experiments).  

\subsection{Initializing the weights}
We randomly initialize the weight matrix $\Phi$ following a zero-mean Gaussian distribution with standard deviation $\sigma$ as:
\begin{equation}
    \Phi \sim \mathcal{N}(0,\sigma)
    \label{iniit}
\end{equation}
Let $\lambda_1 = 0$ in (\ref{ista}), the following equation is obtained by recursion: 
\begin{equation}
    \Bar{c}^{k} = (I_M - \eta_1 \Phi^T \Phi)^{k} \Bar{c}^{0} + \eta_1 \sum_{p=0}^{k-1} (I_M - \eta_1 \Phi^T \Phi)^p \Phi^T \Bar{s}^{\Tilde{p}}
    \label{seri1}
\end{equation}
where $\Tilde{p} = k - 1- p$. A sufficient condition for the convergence of (\ref{seri1}) is to ensure that the \textit{spectral radius} of $(I_M - \eta_1 \Phi^T \Phi)$ is smaller than 1, leading to:
\begin{equation}
    |1 - \eta_1 N\sigma^2| < 1 \Rightarrow \sigma < \sqrt{\frac{2}{\eta_1 N}}
    \label{condw}
\end{equation}

It must be noted that the weight initialization condition (\ref{condw}) naturally holds when $\lambda_1 > 0$ in (\ref{ista}) since the $l_1$ regularization in (\ref{lassodef}) helps shrinking the elements of $\Bar{c}^{k}$ along the iterations which, compared to (\ref{seri1}), leads to a more stable process.


\subsection{Terminating the STDP training}
Detecting when STDP training starts to over-fit and therefore, when it must be stopped, is a key issue rarely mentioned in previous STDP work \cite{8989987}. As in Section \ref{weargue}, we argue that this omission is due to the lack of a link between the SNN and a clearly formulated objective function describing the quantity that the SNN-STDP system seeks to optimize. In the case of our work, we exploit the activity of the \textit{error} layer (Fig. \ref{fig1} b) encoding the re-projection error (\ref{errorwiring}) of the learning objective (\ref{dicodef}) as a bipolar spike train vector:
\begin{equation}
    \begin{cases} \Bar{e}^p(t) = LIF \{\Bar{J}_{in}^e \} \\ 
    \Bar{e}^n(t) = LIF \{-\Bar{J}_{in}^e \} \end{cases}
    \label{reprojvec2}
\end{equation}
 
 We consider the error metric:
\begin{equation}
    \mathcal{L}_{\text{in}} = ||r\{ \Bar{e}^p(t) - \Bar{e}^n(t) \}||_2
    \label{innerlosseq}
\end{equation}
as a measure of the model fitting performance, with $r\{.\}$ denoting the average spiking rate and $\mathcal{L}_{\text{in}}$ referred to as the \textit{inner loss} of the SNN-STDP model. 

\begin{figure}[htbp]
\centering
    \includegraphics[scale = 0.7]{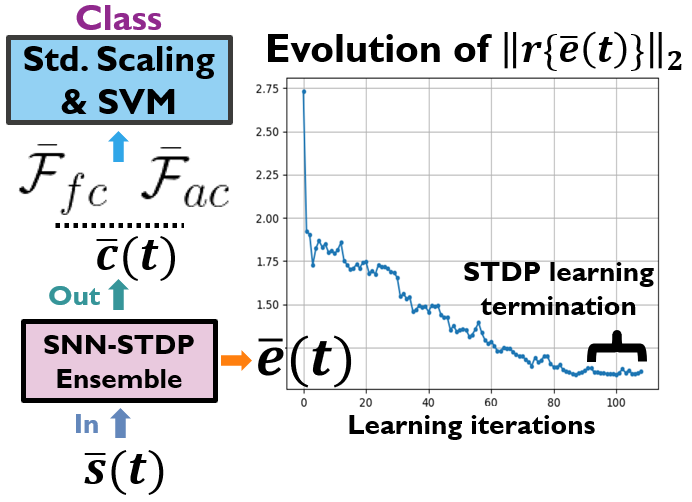}
    \caption{\textit{\textbf{SNN-STDP system architecture} for both the global and the action descriptor. The event image is flattened as $\bar{s}(t)$} and fed to the SNN-STDP ensemble. The SNN output $\bar{c}$ is transformed using (\ref{meanrate}) or (\ref{action_descr}) into feature vectors $\mathcal{\bar{F}}_{fc}$ or $\mathcal{\bar{F}}_{ac}$. During STDP learning, the activity of the error neurons $\bar{e}(t)$ is tracked as $||r\{\bar{e}(t)\}||_2$ and used to terminate training (obtained during training on the IBM DVS128 Gesture dataset in this example).}
    \label{innerloss}
\end{figure}

Following the availability of the inner loss, intrinsically encoded by the error neurons of our SNN system, we can now follow a conventional cross-validation approach in order to detect when STDP training must be terminated (i.e., when the training process is starting to over-fit). In practice, we use a small subset of the training data ($\sim 10$ examples) as the \textit{validation} set that we periodically use to measure the validation inner loss $\mathcal{L}_{\text{in}}$ of the model (e.g., by periodically cycling between 9 training examples for 1 validation pass). As an illustration, Fig. \ref{innerloss} shows the inner loss evolution during the training of our SNN-STDP system on the IBM DVS128 Gesture dataset \cite{8100264}.
We stop the STDP training process when:
\begin{equation}
    \frac{1}{n_\epsilon}\sum_{k=e-n_\epsilon}^{e} |\mathcal{L}_{\text{in}}^{k} - \mathcal{L}_{\text{in}}^{k-1}| < \epsilon
    \label{stop}
\end{equation}
where $e$ denotes the current epoch, $k$ is the epoch index, $n_\epsilon \sim 10$ is the number of epochs along which the inner loss convergence is computed and $\epsilon \sim 10^{-3}$ is the termination threshold.

\subsection{Neural parameter tuning}
\label{akaikepurs}
Choosing the neuron membrane time constant $\tau_m$ and threshold $\mu$ is a typical problem faced during the design of any SNN-STDP learning system. Often, neural parameters are found using lengthy brute force procedures such as manual parameter search or automated grid search, requiring many iterations between the training of the whole system (SNN feature descriptor followed by e.g., an SVM classifier) with the new neural parameters followed by the validation of the resulting model \cite{8879613}. Such a brute force approach is clearly too time-consuming, making it unsuited for e.g., power- and latency-sensitive \textit{edge learning}.

Unlike previous work, our neural parameter search method first uses our derived constraint between $\tau_m$ and $\mu$ (\ref{cond1}) in order to reduce the tuning problem into a uni-dimensional search along the neural threshold $\mu$ only. In addition, we avoid the iterative retraining and validation of the learning system by estimating $\mu$ within a \textit{maximum a posteriori} approach based on the corrected Akaike information criterion $\mathcal{AIC}_c$, exploiting the spiking activity of the error layer in our network (Fig. \ref{fig1} b). The $\mathcal{AIC}_c$ is given by \cite{1100705}:
\begin{equation}
    \mathcal{AIC}_c = -2 \ln \mathcal{P}_{\Theta(\mu)}(\Bar{S}_l, \Phi) + 2 \Theta(\mu) + \frac{2\Theta^2 + 2\Theta}{N - \Theta - 1}
    \label{aiceq}
\end{equation}
with $\mathcal{P}_{\Theta(\mu)}(\Bar{S}, \Phi)$ the likelihood hypothesis function, $\Bar{S}_l$ an input vector from the dataset, $\Phi$ the parameters of the model (i.e., the SNN weights), $N$ the dimension of the input vector and $\Theta(\mu)$ the number of non-zero elements in the output average spike rate vector $r\{ \Bar{c} \}$ (where $r\{.\}$ denotes the rate, the higher $\mu$, the smaller $\Theta$). As likelihood for the SNN in the rate domain, we choose a typical Gaussian model \cite{1100705}:
\begin{equation}
    \mathcal{P}_{\Theta(\mu)}(\Bar{S}_l, \Phi) \sim \exp{(-\frac{1}{2} \frac{||r\{ \Bar{e}_l(t) \}||_2^2}{ \sigma_z^2})}
    \label{modelgaus}
\end{equation}
with $r\{ \Bar{e}_l(t) \}$ denoting the re-projection error in the \textit{spike-rate} domain. In order to tune $\tau_m$ and $\mu$ in practice, we compute the likelihood (\ref{modelgaus}) using a small number ($N_{\text{mini}} \sim 10$) of examples from the training set. We compute the maximum likelihood estimate of the noise variance in the error spike rate vector by computing the variance of $r\{\Bar{e}_l(t) \}, \forall l \in N_{\text{mini}}$ for the limit of $\mu \to 0$ (and hence, $\lambda_1 \to 0$), which reduces (\ref{lassodef}) to the \textit{ordinary least-squares} (OLS) fitting.  
\begin{figure}[htbp]
\centering
    \includegraphics[scale = 0.55]{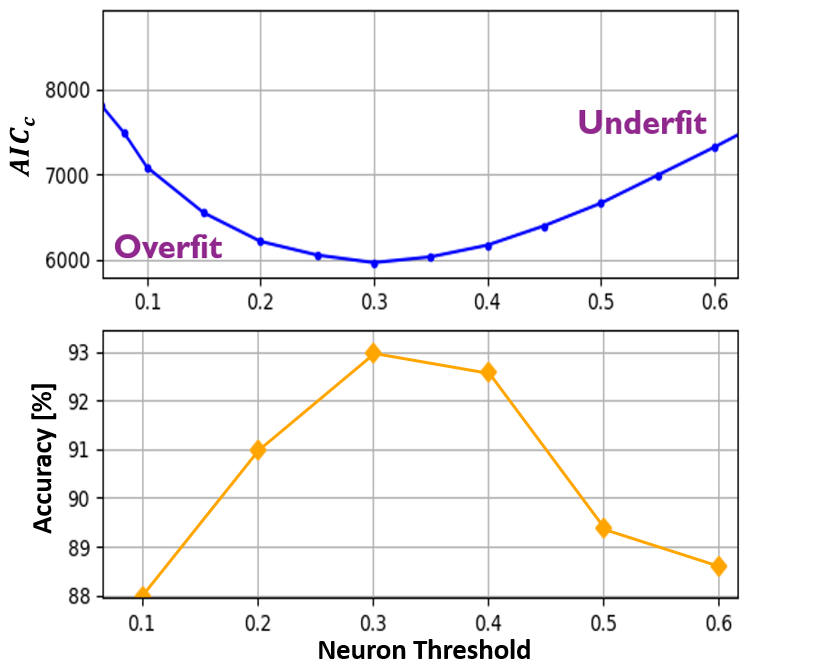}
    \caption{\textit{\textbf{Relationship between $\mathcal{AIC}_c$ (\ref{aiceq}) and the system classification accuracy} (SNN-STDP and linear SVM readout). Assessed on the N-MNIST dataset \cite{10.3389/fnins.2015.00437} using $M=100$ coding neurons. When the neuron threshold $\mu$ is too small, the output of the SNN overfits, leading to a large number of non-zero coefficients for the coding of the input (and vice versa for the underfit region).}}
    \label{aic}
\end{figure}

Fig. \ref{aic} shows the relation between $\mathcal{AIC}_c$ and the accuracy of the classification system obtained by using a linear SVM. As expected, the minimum-AIC threshold $\hat{\mu}$ gives a good indication of the region where an accuracy-maximising $\mu$ can be found. The membrane time constant $\tau_m$ is then found using the proximal constraint (\ref{cond1}).

\section{SNN-STDP system as feature descriptor}
\label{biofeat}
\begin{figure}[!t]
\centering
    \includegraphics[scale = 0.545]{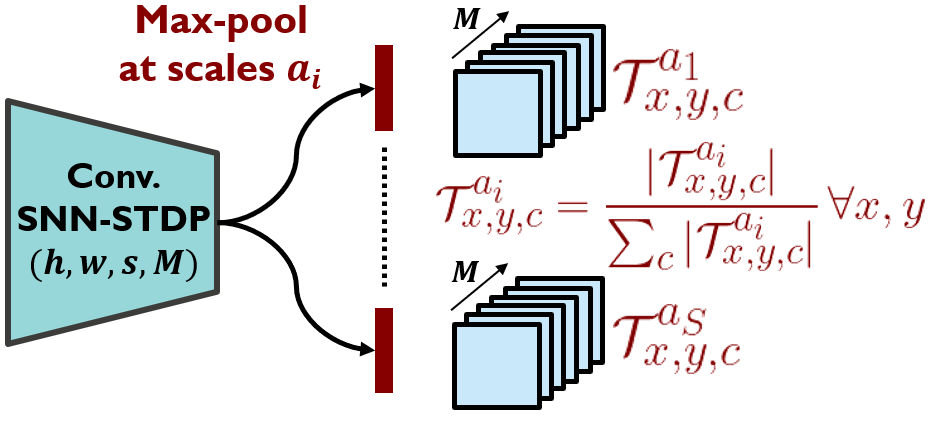}
    \caption{\textit{\textbf{Convolutional SNN-STDP system.} The output feature tensor is max-pooled at different scales $a_i$ following SPP. The obtained bag of tensors is normalized to PDFs along the channel dimension using (\ref{normalic}). Similar to Fig. \ref{innerloss}, we track $||r\{\bar{e}(t)\}||_2$ to terminate STDP during training. As in Fig. \ref{innerloss}, standard scaling and a linear SVM are used to classify the flatten bag of normalized tensors.}}
    \label{convarch}
\end{figure}
This section describes how our SNN-STDP system can be leveraged to build various feature descriptors used to encode the input event data into the learned feature space where linear separability between data classes is enhanced. A standard classifier such as a linear \textit{Support Vector Machine} (SVM) can then be used for classifying the feature vectors at the output of the SNN. We also conduct a power consumption analysis of our SNN and compare our power estimation to other STDP architectures proposed in the literature.
\subsection{Global feature descriptor}
\label{fullycon}
Our SNN-STDP framework can be used to learn global feature descriptors for e.g., event camera data by feeding the event data as flattened spiking vectors to the SNN. The spatio-temporal descriptor $\mathcal{\bar{F}}_{fc}$ is obtained during inference by recording the average firing rate of each spiking output $r\{ \Bar{c}(t) \}$  and normalizing the resulting vector to unit norm in order to provide invariance to event density:
\begin{equation}
    \mathcal{\bar{F}}_{fc} = \frac{r\{ \Bar{c}(t) \}}{||r\{ \Bar{c}(t) \}||_2}  
    \label{meanrate}
\end{equation}
\subsection{Action descriptor}
\label{actiondescr}
It is possible to extend the global feature descriptor described in Section \ref{fullycon} for recognising time-varying signals such as macro-actions and gestures performed in front of an event camera. Compared to \textit{saccades} (vibration of the event camera used for e.g., the N-MNIST dataset acquisitions \cite{10.3389/fnins.2015.00437}), macron-actions cause important oscillations in the rate of the output spike trains of the SNN, due to the long-term periodicity of the performed gestures (such as a hand rotation). Therefore, capturing the correlations between the oscillating spiking rate outputs conveys important information about the performed actions. In order to do so, we concatenate (denoted by the $\oplus$ sign) the Pearson correlation matrix $\rho$ (i.e., the normalized covariance matrix) of $r\{ \Bar{c}(t) \}$ with the matrix $P$ obtained by taking the outer product of the output mean rate vector with itself $P = r\{ \Bar{c}(t) \} r\{ \Bar{c}(t) \}^T$ as follows (the correlations in $\rho$ being computed along the time domain, capturing changes in spike rate):

\begin{equation}
    \mathcal{\bar{F}}_{ac} = \frac{\rho_{\text{tri}} \oplus P_{\text{tri}}}{|| \rho_{\text{tri}} \oplus P_{\text{tri}} ||_2} 
    \label{action_descr}
\end{equation}
where the \textit{tri} sign denotes the upper triangular part of the matrix (since both $\rho$ and $P$ are symmetrical matrices).
\subsection{Convolutional feature descriptor}
\label{convdescr}

\begin{figure}[htbp]
\centering
    \includegraphics[scale = 0.6]{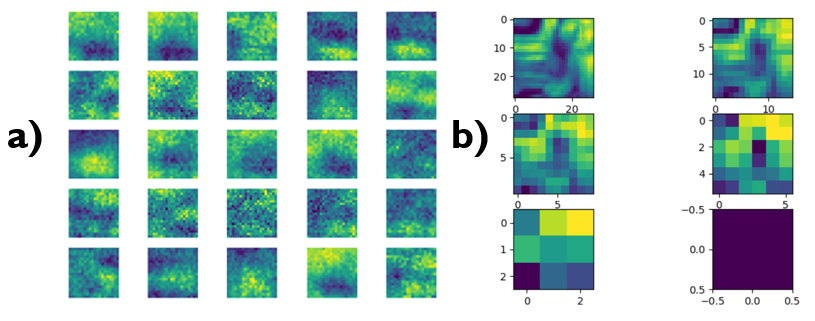}
    \caption{\textit{\textbf{Convolutional kernels and Spatial Pyramidal Pooling (SPP).} a) Subset of convolutional kernels learned on the CIFAR10-DVS dataset \cite{10.3389/fnins.2017.00309}. b) Applying SPP \cite{7005506} to a channel of the output feature map obtained after applying the convolutional SNN-STDP descriptor on an example acquisition from the CIFAR10-DVS dataset \cite{10.3389/fnins.2017.00309}. SPP subsequently applies max-pooling with increasing kernel size.}}
    \label{conk}
\end{figure}

It is naturally possible to extend the SNN-STDP framework described in Section \ref{methodsec} to the case of \textit{convolutional} learning:
\begin{equation}
    \Bar{c}, \Phi = \arg \min_{\Bar{c}, \Phi} \frac{1}{2} ||\Phi * \Bar{c} - \Bar{s}||_{2}^{2} + \lambda_1 ||\Bar{c}||_1 + \frac{\lambda_2}{2} ||\Phi||_F^2
    \label{convvv}
\end{equation}
in order to capture translational invariances \cite{8237828} in the input event data. This is done by sweeping the $h\times w$ visual field of our SNN-STDP system across the event-camera plane with stride $d$ (which we compute in parallel on a GPU during our experiments). Fig. \ref{conk} a) shows a subset of convolutional kernels learned by our SNN-STDP system working under the convolutional setting. After applying the convolutional SNN composed of $M$ kernels to a $H\times W$ image, we obtain a $l_H \times l_W \times M$ tensor $\mathcal{T}_{x,y,c}$ of output spike rates (with $l_H=(H-h)/d + 1$ and $l_W=(W-w)/d + 1$). Then, multi-scale features are captured by max-pooling (in absolute value sense) the tensor $\mathcal{T}_{x,y,c}$ at different scales $a_i$ using spatial pyramidal pooling (SPP) \cite{7005506} (see Fig. \ref{conk} b). 
After applying SPP, we obtain a bag of tensors $\{\mathcal{T}_{x,y,c}^{a_1}, \mathcal{T}_{x,y,c}^{a_2},... \}$. Then, we normalize each tensor along the filter bank channels $c=1,...,M$ in order to provide invariance to event density. Normalization is done by considering the spike rate activity along the kernel dimension $c$ for each pixel $(x,y)$ to represent a feature histogram captured by the SNN-STDP system \cite{8723171}. Therefore, the histograms are normalized to probability density functions as follows:
\begin{equation}
    \mathcal{T}_{x,y,c}^{a_i} \xleftarrow{} \frac{|\mathcal{T}_{x,y,c}^{a_i}|}{\sum_{c} |\mathcal{T}_{x,y,c}^{a_i}|} \hspace{5pt} \forall x,y
    \label{normalic}
\end{equation}

Finally, we aggregate the final feature descriptor $\bar{\mathcal{F}}_{cv}$ by first flattening each normalized tensor $\mathcal{T}_{x,y,c}^{a_i} \hspace{3pt} \forall i$ and then, concatenating the flattened tensors side by side. 

In Section \ref{experres}, we assess the SNN-STDP feature descriptors described above on three benchmark event-based camera datasets (N-MNIST \cite{10.3389/fnins.2015.00437}, CIFAR10-DVS \cite{10.3389/fnins.2017.00309} and IBM DVS128 Gesture \cite{8100264}) and compare our results to the state of the art.

\section{Results}
\label{experres}
In this section, we assess the performance of our proposed SNN-STDP theory and the feature descriptors of Section \ref{biofeat}. Similar to previous works, we consistently use standard scaling followed by a linear SVM classifier \cite{JMLR:v12:pedregosa11a} to evaluate linear separability in the output space. 
\subsection{N-MNIST classification}
\label{nmnistse}
The N-MNIST dataset \cite{10.3389/fnins.2015.00437} features standard MNIST images, re-captured using an event camera. Each acquisition contains a tensor of spike trains (along time and image space). In order to feed the events to our SNN (as a \textit{global} descriptor, see Section \ref{fullycon}), the spiking matrices are flattened to 1-dimensional vectors of spike trains (noted $\Bar{s}(t)$ in Fig. \ref{fig1}). As the N-MNIST matrices are of dimension $34 \times 34$, the input dimension is $N=1156$. We use $M=4000$ coding neurons. The SNN-STDP dictionary is learnt by using the first $40$ training examples of each class, leading to only $N_{\text{ex}} = 400$ training examples in total (see Fig. \ref{badgood}).

After the Akaike-based pursuit of $\hat{\mu}$ (defining $\lambda_1$), we apply a linear SVM classifier on the complete \textit{SNN-encoded} training set. We asses the accuracy of our system on the provided test set. Table \ref{paramsdvstrain} reports the training parameters. Table \ref{bigtable2} reports a superior N-MNIST accuracy for our system compared to state-of-the-art DVS feature descriptors.
\begin{table}[!t]
\centering
\begin{tabular}{|c|c|c|c|c|c|c|c|}
\hline
$\eta_1|\eta_2$ & $\lambda_2$ & $A_+$ & $A_-$ & $\tau_+$ [s] & $\tau_-$ [s] & $\tau_s$ [s]\\
\hline
$1|0.003$ & $0.002$ & 1 & 0.8 & 0.0208 & 0.008 & 0.01\\
\hline
\end{tabular}
\caption{\textit{\textbf{Learning parameters.} The simulation time step is $0.005$s.}}
\label{paramsdvstrain}
\end{table}
\begin{table}[!t]
\begin{tabularx}{0.47\textwidth}{@{}l*{1}{c}c@{}}
\toprule
Architecture  & \hspace{15pt} N-MNIST \% & \hspace{15pt} CIFAR10-DVS \% \\ 
\midrule
MuST (STDP) \cite{8989987}      & \hspace{15pt}89.96       & \hspace{15pt}-          \\ 
HATS \cite{8578284}        & \hspace{15pt}99.1       & \hspace{15pt}52.4         \\ 
DART \cite{8723171}        & \hspace{15pt}97.95       & \hspace{15pt}65.78         \\
Kostadinov et al. \cite{9412631}  &  \hspace{15pt}98.1 & \hspace{15pt} -  \\
\textbf{Ours} (STDP)  &  \hspace{15pt} \textbf{99.26} & \hspace{15pt} $\boldsymbol{73.98\pm0.8}$  \\
\bottomrule
\end{tabularx}
\caption{\textit{\textbf{N-MNIST and CIFAR10-DVS performance.} On N-MNIST, our proposed system is the top performer, with close to $10\%$ gain vs. the state-of-the-art STDP system termed MuST \cite{8989987}. On CIFAR10-DVS, our system outperforms DART by $8.2\%$.}
}
\label{bigtable2}
\end{table}
\begin{table}
  \centering
  \begin{tabular}{@{}lc@{}}
    \toprule
    Architecture & IBM DVS128 \\
    \midrule
    Tempotron (STDP) \cite{chin3} & 60.37   \\
    Reservoir (STDP) \cite{9206681} & 65 \\
    SNN-SBP-STDP \cite{doi:10.1126/sciadv.abh0146} & 84.76\\
    \textbf{Ours} (STDP) & $\boldsymbol{92.5}$  \\
    \midrule
    LSTM \cite{8659288} & 88.17 \\
    cNet \cite{9207109} & 90.46 \\
    CNN \cite{8100264} & 91.77 \\
    PointNet++ \cite{8659288}  & 96.34  \\
    STS-ResNet \cite{DBLP:journals/corr/abs-2003-12346}  & 96.7  \\
    CNN-SRNN \cite{naturemachinein}  & \underline{97.91}  \\
    \bottomrule
  \end{tabular}
  \caption{\textit{\textbf{11-class IBM DVS128 performance.} Our SNN-STDP system outperforms the sate-of-the-art STDP-only system proposed in \cite{9206681} by $27.5\%$ and the STDP-backprop system of \cite{doi:10.1126/sciadv.abh0146} by $7.74\%$. In \cite{9206681}, the authors argue that a source of inefficiency in their system is the fact that STDP training leads to a significant number of noisy, feature-less neurons. In contrast, we did not observe such effect using our proposed SNN-STDP framework. Even though at the advantage of a deep learning system (bottom of the table), our shallow 1-hidden-layer descriptor outperforms the deep 16-layer CNN of \cite{8100264} by 0.73\% 
  and achieves an accuracy within 5.41\% of the state-of-the-art deep CNN-SRNN of \cite{naturemachinein} trained through backprop. This is remarkable given the drastically smaller size of our system and its ability for ultra-low-power STDP learning in edge devices (see Section \ref{powercon} for additional discussions on power consumption).}}
  \label{bigtable3}
\end{table}

\subsection{CIFAR10-DVS classification}
\label{cifar10dvs}
The CIFAR10-DVS dataset \cite{10.3389/fnins.2017.00309} features 10000 standard CIFAR10 images (1000 per class) re-captured using an event camera at a resolution of $128 \times 128$. We use our proposed SNN-STDP system in the \textit{convolutional} setting described in Section \ref{convdescr} with a $20\times 20$ visual field ($h = w = 20$), a stride $d=10$ and $M=256$ coding neurons (found empirically). We use the same learning parameter as in Table \ref{paramsdvstrain}. For SPP at the output of the SNN (see Section \ref{convdescr}), we use a collection of $(2\times2)$, $(3\times3)$, $(6\times6)$ and $(12\times12)$ max-pooling layers. We train the SNN-STDP system using the first 50 training examples of each class to learn a filter bank capturing features such as edges, corners and so on (see Fig. \ref{conk} a). 

After the \textit{a posteriori} estimation of $\hat{\mu}$ using our Akaike-based pursuit, we follow the standard methodology in \cite{8723171} where we randomly keep $90\%$ of the dataset as the training set and $10\%$ for testing. We repeat this process 10 times with different dataset samplings and report the mean accuracy and standard deviation in Table \ref{bigtable2} (+8.2\% vs. \cite{8723171}).   
\subsection{IBM DVS128 Gesture classification}
Traditionally, event-based descriptors have mostly been assessed for image recognition tasks, and less so for action recognition, which has mostly been tackled by end-to-end deep learning systems (see Table \ref{bigtable3}). We argue that it is equally important to assess spatio-temporal descriptors on tasks such as gesture recognition, which better corresponds to real-world applications of event cameras compared to datasets capturing still images with vibrating cameras (see \cite{DBLP:journals/corr/abs-1807-01013} for a comprehensive discussion). To this end, we assess our system as an \textit{action} descriptor (see Section \ref{actiondescr}) with $M=1500$ coding neurons on the IBM DVS128 gesture dataset \cite{8100264} following the train-test procedure of \cite{8100264}. We outperform same-class methods (+7.7\%) and even three deep networks in Table \ref{bigtable3}.

\subsection{Power consumption analysis}
\label{powercon}
In Section \ref{lintro}, we motivated the use of SNN-STDP system for designing ultra-low power networks that can learn with tight power budgets. It is therefore important to provide an estimate of the power consumption during SNN learning and inference.

\begin{figure}[htbp]
\centering
    \includegraphics[scale = 0.5]{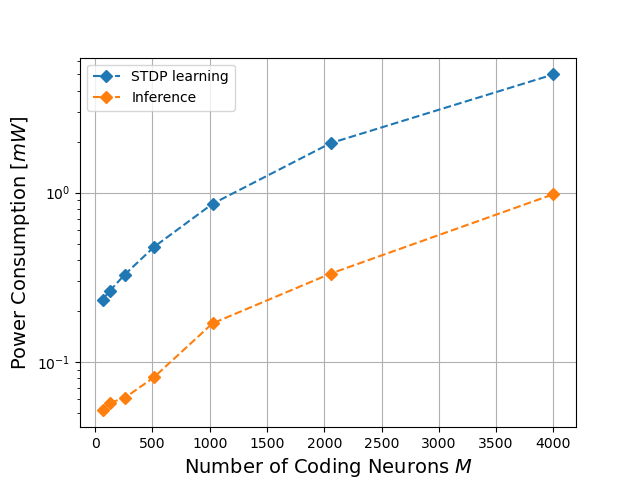}
    \caption{\textit{\textbf{Power consumption} during STDP learning and inference, estimated on the N-MNIST dataset \cite{10.3389/fnins.2015.00437} using hardware metrics provided in \cite{ubrain}.}}
    \label{energy_mnist}
\end{figure}

To this end, we estimate in simulation the power consumption of our SNN-STDP architecture following the hardware metrics of the $\mu$Brain SNN chip \cite{ubrain} (fully synthesizable state-of-the-art architecture specially tailored for ultra-low power applications):
\begin{multline}
     P_c = \frac{1}{T_p} \times (N_{spikes} \times E_{dyn} + T_p \times P_{stat}\\
    + N_{read} \times E_{read} + N_{write} \times E_{write})  
\end{multline}
where $N_{spikes}$ is the total number of spikes during training or inference, $E_{dyn}= 2\times 2.1$ pJ is the energy per spike (taking into account the broadcasting of the spike to the next layer), $T_p$ is the time duration of the training or inference process, $P_{stat} = 2 \times 73$ $\mu$W is the static leakage, $N_{read}$ and $N_{write}$ are respectively the number of SRAM fetch and store operations and $E_{read}$ and $E_{write}$ are the respectively the energy per SRAM fetch or store. We use a (pessimistic) margin of $\times 2$ for $P_{stat}$ and $E_{dyn}$ in order to take into account the STDP circuit overhead (not implemented in \cite{ubrain}). Fig. \ref{energy_mnist} shows the estimated power consumption in function of the number of coding neurons $M$ for STDP training and inference on an input signal $\bar{s}$ of dimension $N=1156$ (e.g., N-MNIST dataset). Table \ref{tablepowr} shows that our power consumption estimate is in the same order of magnitude than previously-proposed SNN-STDP circuits (input dimension $N$ kept the same during this comparison), while being the top performer among SNN-STDP processing architectures in terms of accuracy (see Table \ref{bigtable2} and \ref{bigtable3}).  

\begin{table}[htbp]
\begin{center}
\begin{tabular}{|c|c|c|c|c|c|}
\hline
  & \textbf{Ours} & \cite{stdpchip1} & \cite{stdpchip2} & \cite{fedostdp}\\
\hline
 STDP [mW]  & 5 & 3.348 & 9.4 &  4\\
 \hline
 Inference [mW]  & 1 & $<3.348$ & 6.2 & $<4$\\
 \hline

\end{tabular}
\end{center}
\caption{\textit{\textbf{SNN-STDP power consumption}. Input dimension $N=1156$ for all power estimates (typical MNIST-like size used for power benchmarking in literature \cite{stdpchip1}).}}
\label{tablepowr}
\end{table}

\section{Conclusion}
\label{concsec}
This paper has presented an optimization-based theoretical framework explaining how biologically-plausible SNN-STDP ensembles can perform unsupervised feature extraction by linking their dynamics with the theory of joint dictionary learning and LASSO basis pursuit. In addition, theoretical foundations have been provided for tuning the network parameters without extensive empirical parameter search. Our methods have been successfully applied on several DVS camera datasets, reporting competitive performance against the state of the art (+8.2\% on CIFAR10-DVS compared to conventional descriptors, +9.3\% on N-MNIST and +7.74\% on IBM DVS128 Gesture compared to state-of-the-art STDP systems). A power consumption analysis of our system has also been provided which motivates the use of our SNN-STDP methods for ultra-low-power learning and inference in edge devices. Finally, our framework contributes towards an optimization-based and biologically-realistic theory of cortical vision. 

\section*{Acknowledgement}
We would like to thank Dr. Lars Keuninckx for the helpful discussion. The research leading to these results has received funding from the Flemish Government (AI Research Program) and the European Union's ECSEL Joint Undertaking under grant agreement n° 826655 - project TEMPO.

%

\begin{IEEEbiography}[{\includegraphics[width=1in,height=1.5in,clip,keepaspectratio]{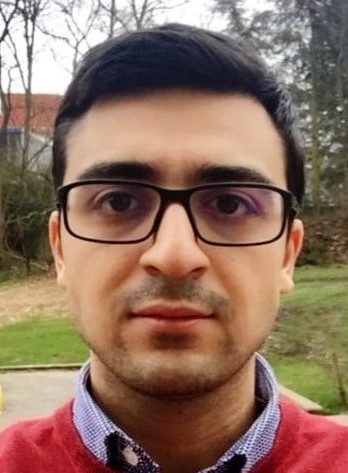}}]{Ali Safa}
(Student Member, IEEE) received the MSc degree (summa cum laude) in Electrical Engineering from the Universit\'e Libre de Bruxelles, Belgium. He joined IMEC and the Katholieke Universiteit Leuven (KU Leuven), Belgium in 2020 where he is currently working toward the PhD degree on AI-driven processing and sensor fusion for extreme edge applications.  
\end{IEEEbiography}

\begin{IEEEbiography}[{\includegraphics[width=1in,height=10in,clip,keepaspectratio]{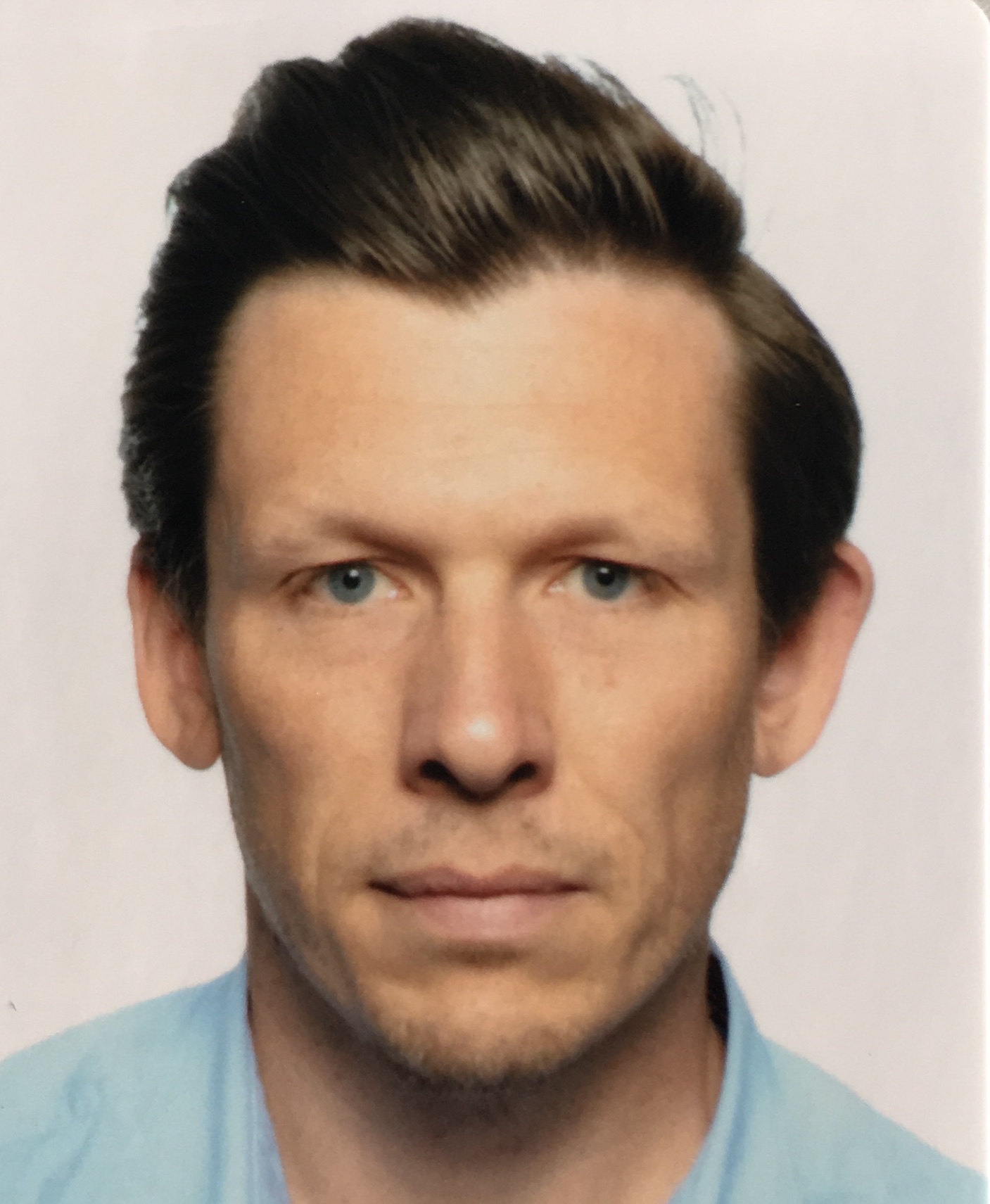}}]{Ilja Ocket}
(Member, IEEE) received the MSc and the PhD degrees in Electrical Engineering from KU Leuven, Leuven, Belgium, in 1998 and 2009, respectively. He currently serves as program manager for neuromorphic sensor fusion at the IoT department of imec, Leuven, Belgium. His research interests include all aspects of heterogeneous integration of highly miniaturized millimeter wave systems, spanning design, technology and metrology. He is also involved in research on using broadband impedance sensing and dielectrophoretic actuation for lab-on-chip applications.

\end{IEEEbiography}

\begin{IEEEbiography}[{\includegraphics[width=1in,height=10in,clip,keepaspectratio]{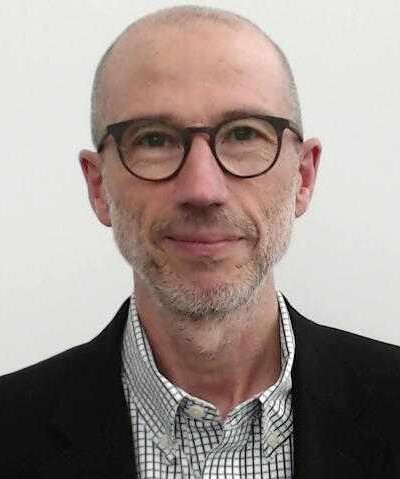}}]{Andr\'e Bourdoux}
(Senior Member, IEEE)
received the M.Sc. degree
in electrical engineering from the Universit\'e Catholique de Louvain-la-Neuve, Belgium,
in 1982. In 1998, he joined IMEC, where he is currently a Principal Member of Technical Staff with
the Internet-of-Things Research Group. He is a
system-level and signal processing expert for both
the mm-wave wireless communications and radar
teams. He has more than 15 years of research experience in radar systems and 15 years of research
experience in broadband wireless communications. He holds several patents
in these fields. He has authored or coauthored over 160 publications in books
and peer-reviewed journals and conferences. His research interests include
advanced signal processing, and machine learning for wireless physical layer
and high-resolution 3D/4D radars.
\end{IEEEbiography}

\begin{IEEEbiography}[{\includegraphics[width=1in,height=10in,clip,keepaspectratio]{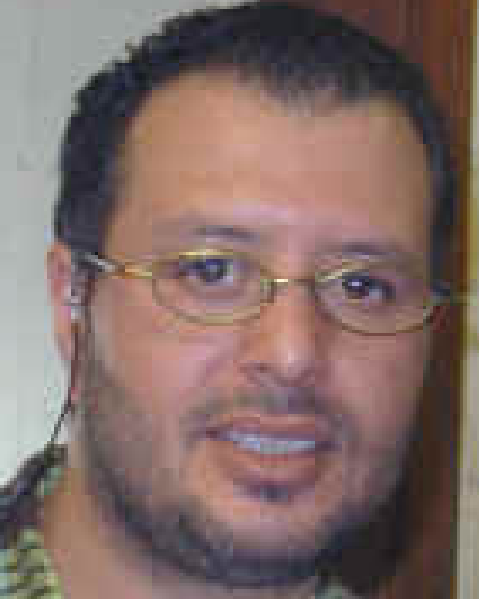}}]{Hichem Sahli}
received the degree in mathematics and computer science, the D.E.A. degree in
computer vision, and the Ph.D. degree in computer
sciences from the Ecole Nationale Superieure de
Physique Strasbourg, France. Since 2000, he has
been a Professor with the Department of Electronics and Informatics (ETRO) and a Scientist with
the Interuniversitair Micro-Elektronica Centrum
VZW (IMEC). He coordinates the Audio-Visual
Signal Processing Laboratory (AVSP) within
ETRO. AVSP conducts research on applied and theoretical problems related
to machine learning, signal and image processing, and computer vision. The
group explores and capitalizes on the correlation between speech and video
data for computational intelligence where efficient numerical methods of
computational engineering are combined with the problems of information
processing.
\end{IEEEbiography}

\begin{IEEEbiography}[{\includegraphics[width=1.1in,height=1.4in,clip,keepaspectratio]{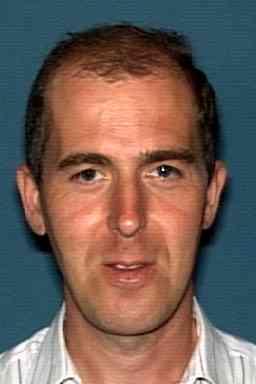}}]{Francky Catthoor}
(Fellow, IEEE) received a Ph.D. in electrical engineering from KU Leuven,
Belgium in 1987.  Between 1987 and 2000, he has headed several research
domains in the area of synthesis techniques and architectural methodologies.
Since 2000 he is strongly involved in other activities at IMEC including
co-exploration of application, computer architecture and deep
submicron technology aspects, biomedical systems and IoT sensor nodes,
and photo-voltaic modules combined with renewable energy systems,
all at IMEC Leuven,  Belgium. Currently, he is an IMEC senior fellow.
He is also part-time full professor at the Electrical Engineering department of the KU Leuven (ESAT).
He has been associate editor for several IEEE and ACM journals and was elected IEEE fellow in 2005.

\end{IEEEbiography}

\begin{IEEEbiography}[{\includegraphics[width=1.1in,height=1.4in,clip,keepaspectratio]{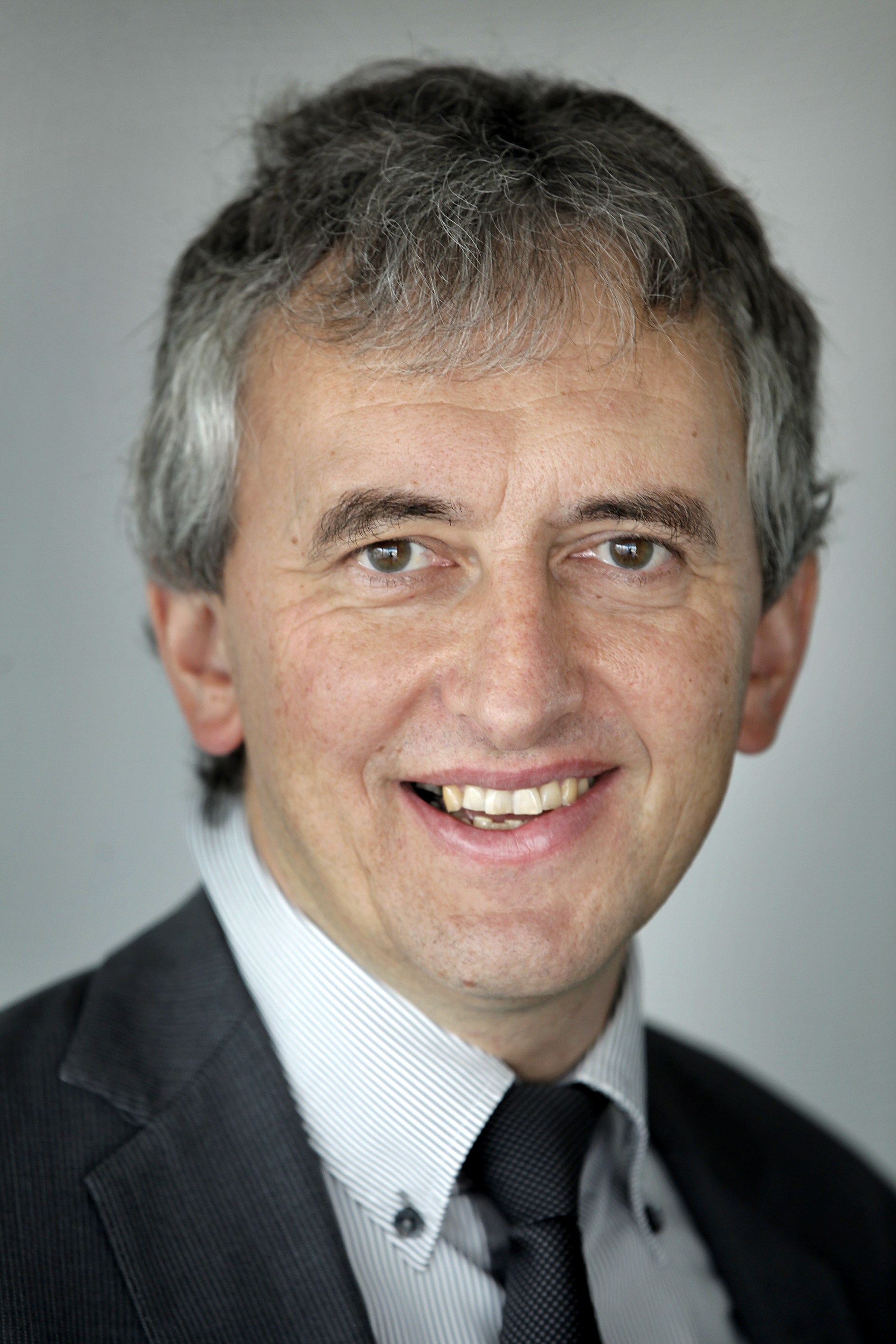}}]{Georges G.E Gielen}
(Fellow, IEEE) received the MSc and PhD degrees in Electrical Engineering from the Katholieke Universiteit Leuven (KU Leuven), Belgium, in 1986 and 1990, respectively. He currently is Full Professor in the MICAS research division at the Department of Electrical Engineering (ESAT) at KU Leuven. Since 2020 he is Chair of the Department of Electrical Engineering. His research interests are in the design of analog and mixed-signal integrated circuits, and especially in analog and mixed-signal CAD tools and design automation. He is a frequently invited speaker/lecturer and coordinator/partner of several (industrial) research projects in this area, including several European projects. He has (co-)authored 10 books and more than 600 papers in edited books, international journals and conference proceedings. He is a 1997 Laureate of the Belgian Royal Academy of Sciences, Literature and Arts in the discipline of Engineering. He is Fellow of the IEEE since 2002, and received the IEEE CAS Mac Van Valkenburg award in 2015 and the IEEE CAS Charles Desoer award in 2020. He is an elected member of the Academia Europaea.
\end{IEEEbiography}








\end{document}